\def\year{2020}\relax
\documentclass[letterpaper]{article} 
\usepackage{aaai20}  
\usepackage{times}  
\usepackage{helvet} 
\usepackage{courier}  
\usepackage[hyphens]{url}  
\usepackage{graphicx} 
\usepackage{graphicx}  
\usepackage{amsmath}
\usepackage{algorithm}
\usepackage{algorithmic}
\usepackage{amssymb}
\usepackage{multirow}
\usepackage[misc]{ifsym}
\usepackage{makecell}
\title{All You Need Is Boundary: Toward Arbitrary-Shaped Text Spotting}
\author{
Hao Wang\textsuperscript{\rm 1}\thanks{Authors contribute equally.},
Pu Lu\textsuperscript{\rm 1}\footnotemark[\value{footnote}],
Hui Zhang\textsuperscript{\rm 1}\footnotemark[\value{footnote}],
Mingkun Yang\textsuperscript{\rm 1},
Xiang Bai\textsuperscript{\rm 1}\thanks{Corresponding authors.},\\
\bf \Large
Yongchao Xu\textsuperscript{\rm 1},
Mengchao He\textsuperscript{\rm 2},
Yongpan Wang\textsuperscript{\rm 2},
Wenyu Liu\textsuperscript{\rm 1}\footnotemark[\value{footnote}]\\
\textsuperscript{\rm 1}Huazhong University of Science and Technology,
\textsuperscript{\rm 2}Alibaba Group\\
\{wanghao4659,lupu,huizhang0110,yangmingkun,xbai,yongchaoxu,liuwy\}@hust.edu.cn\\
mengchao.hmc@alibaba-inc.com, yongpan@taobao.com
}

\begin{document}

\maketitle

\begin{abstract}
Recently, end-to-end text spotting that aims to detect and recognize text from cluttered images simultaneously has received particularly growing interest in computer vision. Different from the existing approaches that formulate text detection as bounding box extraction or instance segmentation, we localize a set of points on the boundary of each text instance. With the representation of such boundary points, we establish a simple yet effective scheme for end-to-end text spotting, which can read the text of arbitrary shapes. 
Experiments on three challenging datasets, including ICDAR2015, TotalText and COCO-Text demonstrate that the proposed method consistently surpasses the state-of-the-art in both scene text detection and end-to-end text recognition tasks.

\end{abstract}

\section{Introduction}
Automatic reading text from natural images has attracted great attention due to its wide practical applications such as office automation, network content security, intelligent transportation system \cite{2018ZhuTS,rong2016recognizing}, geo-location, and visual search \cite{bai2018integrating}.

In the past decade, scene text detection and recognition are extensively studied as two separated sub-tasks of a reading system, but in fact, text detection and recognition are highly relevant and complementary to each other. This assumption is confirmed by the recent end-to-end text spotting methods \cite{jaderberg2016reading,TextBoxes,FOTS,ShenICCV2017,HeCVPR2018,busta2017deep,MaskTextspotter} that combine the detection and recognition stages with an end-to-end trainable neural network. These spotting methods follow a similar pipeline. First, the horizontal/oriented bounding box of each text instance is detected. Then, the image patches or CNN features inside the detected bounding boxes are cropped and fed to a sequence recognition model. Benefiting from feature sharing and joint optimization, the performances of detection and end-to-end recognition can be enhanced at the same time. 

\begin{figure}[t]
  \includegraphics[width=0.95\linewidth]{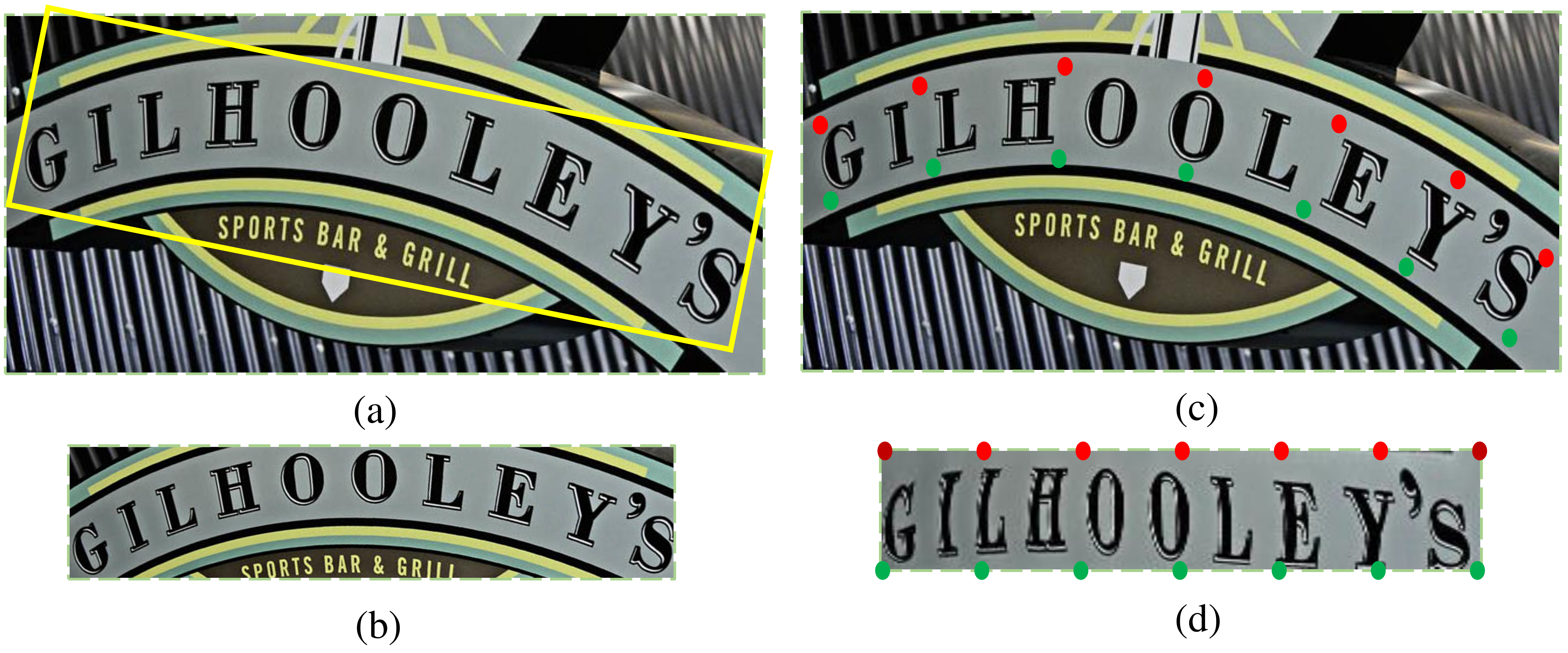}
  \caption{
Illustrations of two kinds of methods for text region representation.  (a) An oriented rectangle box is used to represent the text region and cropped as (b); (c) A set of boundary points are used to represent the text region, and it can be transformed into a horizontal region like (d).}
  \label{abstract}
\end{figure}

Despite promising progress, most existing spotting methods \cite{TextBoxes,FOTS,ShenICCV2017,HeCVPR2018,busta2017deep,jaderberg2016reading} suffer from dealing with text of irregular shapes, such as curve text. For a general end-to-end OCR system, it is inevitable to handle the text with arbitrary shapes, as curve text and other types of irregular text are very common in our real-world. In \cite{FOTS,ShenICCV2017,HeCVPR2018,busta2017deep,liu2017deep,he2017deep}, the detected bounding box of each text instance is represented with a rectangle, which only can tightly cover straight text instances. Rectangular boxes have high limitations in describing irregular text for an end-to-end text spotter since it more or less contains background information which brings difficulties to the text recognition stage, as shown in Fig.~\ref{abstract}(a). Recently, an end-to-end OCR model \cite{MaskTextspotter} for spotting arbitrary-shaped text is presented based on Mask RCNN, which tackles text detection and recognition via instance segmentation and achieved state-of-the-art result. However, this method needs extra character-level annotations for training, and its processes of instance segmentation brings more computational burden. 

\begin{figure*}[t]
    \includegraphics[width=0.85\linewidth]{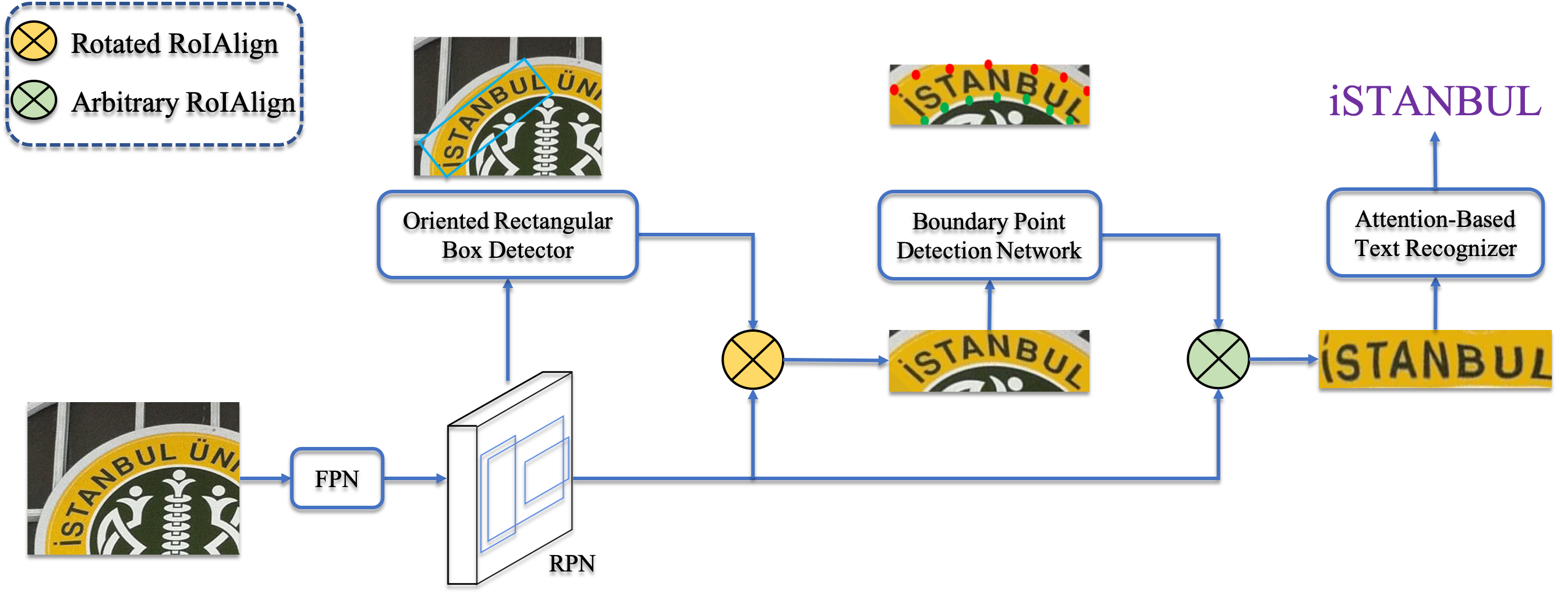}
\centering
  \caption{An overview of the proposed method. The Oriented Rectangular Box Detector is used to predict oriented rectangular box. The Boundary Point Detection Network inputs each rotated Region of Interest (RoI) feature and predicts a set of boundary points. Finally, the Arbitrary RoIAlign is used to get rectified features for attention-based text recognizer to predict text labels. Note that, for all figures, we use the input image for illustration, but operations are actually conducted on feature maps.}
  \label{framework}
\end{figure*}

In this paper, we propose an end-to-end trainable network for spotting arbitrary-shaped text without character-level annotations. Instead of detecting a rectangle bounding box, our detection is performed by localizing the boundary of a text instance. More specifically, the aim of our detection is to predict a set of boundary points, which are more flexible for describing various shapes of scene text, often embodied in two dimensional space, as shown in Fig.~1(c). The usage of boundary points has three advantages for building an end-to-end OCR system: 1) CNN features of irregular text region can be accurately acquired with boundary points, resulting in effectively eliminating the disturb of background noise to the subsequent recognition; 2) With boundary points, irregular text can be easily transformed or rectified into a regular one (i.e. horizontal text), as described in Fig.~1(d), which is realistic input for a sequence recognition model. Similar to a recent model for irregular text recognition \cite{ASTER}, such a transformation operation can be simply implemented and differentiable in CNN; 3) The position of boundary points can be easily refined through back propagation when training a recognition model, fully enjoying the improvement of detection performance from recognition stage. Therefore, boundary points appear to be a reasonable representation that can smoothly and effectively bridge text detection and recognition modules.

However, directly detecting boundary points of text is challenging due to the diversity of text shape and scale. To effectively extract text boundary points, we adopt a coarse-to-fine strategy: First, the minimum oriented rectangular box of each text instance is detected with a two-stage CNN detector, as shown in in Fig.~1(a); Then, the boundary point prediction is performed in the oriented rectangular box.
Our experiments have validated the effectiveness of the proposed boundary point detection. Additionally, benefiting from the representation of boundary points, the proposed method achieves state-of-the-art performance in both text detection and text spotting on several benchmark datasets.

The contributions in this work are two-fold:
1) We recommend the representation of boundary points for end-to-end text spotting, which is more suitable than a rectangular box or segmentation mask for connecting detection and recognition modules.   
2) We design a novel end-to-end trainable network for joint optimizing boundary point detection and text recognition, which can read both straight and curve text.

\begin{figure*}[tp]
\includegraphics[width=0.85\linewidth]{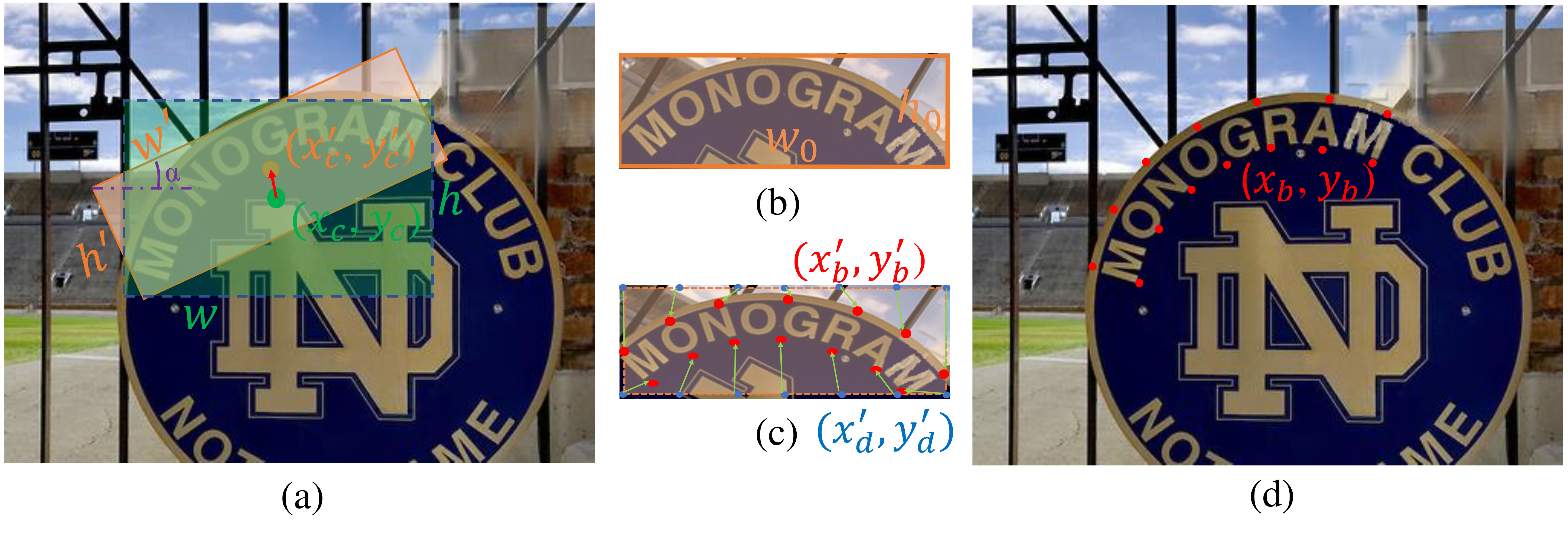}
\centering
  \caption{
  (a) Regression from the axis-aligned box of RPN to the oriented rectangular box. (b) The minimum rectangular box. (c) Regression from a set of default points which are uniformly distributed at the upper and lower sides of the minimum rectangular box to boundary points. (d) Aligning boundary points to the original image.
  }
  \label{regression}
\end{figure*}

\section{Related Work}

Scene text reading has attracted great attention in computer vision. Plenty of excellent works have appeared in the past decade. Due to the page limit, we can not detail the works about scene text detection and recognition which are referred to \cite{long2018scene,zhu2016scene}. Here, related works about scene text spotting are introduced.

Methods on text spotting could be roughly divided into two categories according to the representation of text region: rectangular box based methods and segmentation based methods. In the first category, previous methods \cite{TextBoxes,jaderberg2016reading} train text detector and recognizer separately. And image patches are cropped based on rectangular boxes for recognition. The separate pipeline results in unsatisfactory performance on both tasks, since the relationship within them is ignored. \cite{FOTS,ShenICCV2017,HeCVPR2018,busta2017deep} share a common idea: text region features are extracted for subsequent recognizer. In \cite{ShenICCV2017}, text region is formulated as horizontal rectangular box, which can only support horizontal text instance. Methods in \cite{FOTS,HeCVPR2018,busta2017deep,Sun2018TextNetIT} extract text features in the minimum oriented rectangular box, which can tackle oriented text. But those works suffer from handling curved text.
As the only segmentation based method \cite{MaskTextspotter}, each text instance and corresponding characters are segmented. Although curved text could be represented by the segmentation map, extra character-level annotations are required. In addition, the trained model ignores contextual information between characters of a word, which will affect the recognition performance.

In order to read arbitrary-shaped text in an end-to-end manner, a set of boundary points are proposed to describe text regions. 
Although boundary points are also used for text region representation in \cite{zhang2019look,zhu2018sliding}, a complex post processing is required to generate them, which is not differentiable for training.
In spite of predicting boundary points directly without post processing in \cite{Wang_2019_CVPR}, a recurrent neural network based network is time consuming for end-to-end task.
Compared to \cite{FOTS,ShenICCV2017,HeCVPR2018}, the novel representation is more flexible and accurate to represent text with arbitrary shapes than rectangle box. Compared to \cite{MaskTextspotter}, such a description is free from character-level annotations. More details about the proposed method will be introduced in the following sections.

\section{Methodology}

As illustrated in Fig. \ref{framework}, our pipeline is composed of three parts: the oriented rectangular box detector, the boundary point detection network, and the recognition network. As for the oriented rectangular box detector, we first apply a RPN~\cite{RPN}, where the backbone is FPN~\cite{FPN} equipped with ResNet-50~\cite{Resnet}, to generate horizontal text proposals.
Then, an oriented rectangular box of each proposal is generated via predicting its center point, height, width, and orientation.
Next, the boundary points of each oriented rectangular box are regressed by the boundary point detection network (BPDN). Finally, with the predicted boundary points, the feature maps are rectified as regular ones for the subsequent text recognizer.

As shown in Fig.~\ref{regression}(a), the boundary points can be predicted for each horizontal proposal by BPDN, but we observe that BPDN suffers from text instances of various directions and shapes since such cases contain more background noise and have stronger deformation. To alleviate this effect, an oriented rectangular box of each proposal is predicted, with which the feature maps are transformed into horizontal ones via RotatedRoIAlign \cite{RotatedROIAlign}, as illustrated in Fig.~\ref{regression}(b). Concretely, we follow the method proposed in \cite{RRPN} to obtain the oriented rectangular box by predicting its center point, height, width, and orientation.
Here, a module composed of three stacked fully connected layers is designed. More details about detecting the oriented rectangle are referred to Fig.~\ref{regression}(a). 

\subsection{Boundary Point Detection Network}

BPDN consists of four stacked 3$\times$3 convolutional layers and one fully connected layer. Inspired by RPN where proposals are regressed based on default anchors, a set of default points are predefined for boundary points to refer, as shown in Fig.~\ref{regression}(c). Specifically, $K$ points are equidistantly sampled on each long side of text instance as target boundary points. And corresponding default points are evenly placed along long sides of the minimum rectangular box. Instead of directly predicting the coordinates of boundary points, offsets to its associated default points are first generated. The module predicts a 4$K$-d vector which is coordinate offsets (2-d) of 2$K$ boundary points. Given the coordinate offsets ($\Delta x'$, $\Delta y'$), the boundary point ($x'_{b},y'_{b}$) can be obtained from

\begin{equation}
\begin{aligned}
	&x'_{b} = x'_{d} + w_0\Delta x'\\
	&y'_{b} = y'_{d} + h_0\Delta y',
\end{aligned}
\end{equation}
where ($x'_{d},y'_{d}$) represents default point. $w_{0}$ and $h_{0}$ are the width and height of the minimum rectangular box.

To be consistent with original features, we align the boundary points ($x'_{b},y'_{b}$) in transformed horizontal feature maps to the original ones ($x_{b},y_{b}$) using

\begin{equation}
\label{transform}
\left [ 
\begin{array}{c}
	x_{b} \\
	y_{b} \\
	1
\end{array}
\right ]= M^{-1}\left[ 
\begin{array}{c}
	x'_{b} \\
	y'_{b} \\
	1
\end{array}
\right],
\end{equation}
\begin{equation}
\label{M}
M =\left [
\begin{array}{ccc}
	s_wcos\alpha& -s_hsin\alpha& -s_wx'_{c}cos\alpha+s_hy'_{c}sin\alpha\\
	s_wsin\alpha& s_hcos\alpha& -s_wx'_{c}sin\alpha-s_hy'_{c}cos\alpha\\
	0& 0& 1 \\
\end{array}
\right]\!,
\end{equation}
($x'_{c}, y'_{c}$) is the center point of the oriented rectangle. $s_{w}$ and $s_{h}$ equal to $w_o/w'$ and $h_o/h'$ respectively. And $\alpha$ is the angle from the positive direction of the x-axis to the direction parallel to the long side of the oriented rectangle.

Following \cite{ASTER}, ArbitraryRoIAlign is adopted to flatten features of text instances with arbitrary shapes. Specifically, given the boundary points of each text instance, Thin-Plate-Spline transformation \cite{TPS} is adopted to rectify the features to regular ones.
\subsection{Recognition Network}
 \begin{table}[t]
 \centering
 \begin{tabular}{p{.12\columnwidth} p{.22\columnwidth} l c}
	\hline
 	\multirow{2}*{} & \multirow{2}*{Type} & Configurations  & Out\\
	& & \small{[size,stride,padding]} & Channels \\
	\hline \multirow{6}*{Encoder} & conv\_bn\_relu & [3,1,1] & 256 \\
	 & max-pool & [2,1,0] & 256 \\
	 & conv\_bn\_relu & [3,1,1] & 256 \\
	 & max-pool & [2,1,(0,1)] & 256 \\
	 & conv\_bn\_relu & [3,1,1] & 256 \\
	 & max-pool & [2,1,(0,1)] & 256 \\
	 \hline
	 \multirow{2}*{Decoder}  & Att. GRU & & 256 \\
	 & FC & & $|S|$\\
 \hline
 \end{tabular}
 \caption{
  The architecture of the recognition branch, which consists of three stacked convolutional layers, ``Att. GRU" which stands for attentional GRU decoder and a fully-connection layer. $|S|$ represents the number of decoded characters. We set the number of decoded characters in our experiments to 63, which corresponds to digits (0-9), English characters (a-z/A-Z), and an end-of-sequence symbol.
  }
  \label{recognition}
 \end{table}
CRNN \cite{CRNN} is the first method to treat text recognition as a sequence-to-sequence problem by combining CNN and RNN in an end-to-end network. In some latest works \cite{zhan2019esir,yang2019symmetry,luo2019moran}, recognition network is a common attentional sequence-to-sequence network.
The recognizer predicts a character sequence from the rectified features. The architecture of the recognition branch is given in Tab.~\ref{recognition}. Firstly, the rectified features are fed into encoder to extract higher-level feature sequence $F \in \Re^{n\times{C}}$. Then the attention-based decoder is adopted to translate $F$ into a symbol sequence $y = (y_1,...,y_T)$, where T is the length of the label sequence. At step $t$, the decoder predicts a character based on the encoder output $F$, the internal state $s_{t-1}$ and the result $y_{t-1}$ predicted in the last step. In the current step, the decoder starts by computing a vector of attention weights, $\alpha_{t}$, through its attention mechanism. Then, the weighted feature $g_t$ is calculated according to

\begin{equation}
\label{weightfeature}
\begin{aligned}
	& g_t = \sum_{i=1}^n\alpha_{t,i}F_i\\
	& \alpha_{t,i} = exp(e_{t,i})/\sum_{j=1}^nexp(e_{t,j})\\
	& e_{t,j} = w^Ttanh(Ws_{t-1}+VF_i+b),
\end{aligned}
\end{equation}
where $w$, $W$, $V$ and $b$ are trainable weights.

Taking $s_{t-1}$, $g_t$ and $y_{t-1}$ as inputs, the RNN calculates an output vector $x_t$ and a new state vector $s_t$ via

\begin{equation}
\label{RNN}
	(x_t, s_t) = RNN(s_{t-1}, (g_t, onehot(y_{t-1}))),
\end{equation}
where $(g_t, onehot(y_{t-1}))$ is the concatenation of $g_t$ and the one-hot embedding of $y_{t-1}$. In our method, a GRU is used as RNN unit. Finally, a distribution of the current-step symbol is predicted through
\begin{equation}
\label{Prediction}
	p(y_t) = softmax(W_ox_t + b_o),
\end{equation}
where $W_o$ and $b_o$ are learnable parameters.

\subsection{Loss Functions}
The objective function consists of four parts, which is defined as follows,
\begin{equation}
	L = L_{rpn} + L_{or} + L_{bp} + L_{recog},
\end{equation}
where $L_{rpn}$ is the loss of RPN, which is identical as in \cite{RPN}. $L_{or}$ is the loss that regression from axis-aligned rectangular proposal to oriented rectangular box, which is similar as \cite{RRPN}. Since the above losses are not our main contributions, we do not detail them here due to the page limit. $L_{bp}$ is the loss of boundary point regression which is calculated as Smoothed-L1 loss. The loss function can be formulated as
\begin{equation}
\begin{aligned}
\label{localization}
	L_{bp} = \frac{1}{2K}\sum_{i=1}^{2K}\left({Smooth_{L1}(\hat{x}'_{b,i}, x'_{b,i}) +} \right.\\
	\left.{Smooth_{L1}(\hat{y}'_{b,i}, y'_{b,i})}\right),
\end{aligned}
\end{equation}

\noindent where $(x'_{b,i}, y'_{b,i})$ is the $i$-th predicted boundary point, whose associated target boundary point is $(\hat{x}'_{b,i}, \hat{y}'_{b,i})$.

In the recognition network, the recognition loss can be formulated as
\begin{equation}
	\label{recognitionloss}
	L_{recog}= -\frac{1}{T}\sum_{t=1}^T\log p(y_t).
\end{equation}

\section{Experiments}
To confirm the effectiveness of the proposed method on arbitrary-shaped text spotting, we conduct exhaustive experiments and compare with other state-of-the-art methods on four popular benchmarks which consist of a horizontal text set ICDAR2013 \cite{ICDAR2013}, two oriented text sets ICDAR2015 \cite{ICDAR2015} and COCO-Text \cite{COCOText}, a curved text set TotalText \cite{ch2019total}. The details about these datasets are as follows.

\subsection{Datasets}

\textbf{SynthText} \cite{SynthText} has about 800,000 images, which are generated via synthesizing engine. Text instances within the images are multi-oriented, whose annotations consist of word-level and character-level oriented bounding boxes, as well as text sequences. 

\begin{algorithm}
    \caption{Generate Target Boundary Points}
    \label{alg:A}
    \begin{algorithmic}[1]
        \REQUIRE ~~ \\
        Points on each long side: $P = \{p_0, p_1,..., p_{N-1} \}$.\\
        The excepted number of sampled points: K.
        \ENSURE ~~ \\
        The generated target boundary points:\\
        $Q = \{q_0, q_1,..., q_{K-1} \}$.
        \STATE $distance[0]$ =$0$; $len[0]=0$;
        \FOR{i=0; i $<$ N-1; i++}
        \STATE $//$ calculate distance between two adjacent points
        \STATE $distance[i+1]$ = distance$(p_i, p_{i+1})$;
        \ENDFOR
        \FOR{j=1; j$<$N; j++}
        \STATE $len[j] = len[j-1]+distance[j] $;
        \ENDFOR
        \STATE $average\_distance = \frac{1}{K-1}*\sum{distance[i]}$;
        \FOR{i=0; i $<$ K; i++}
        \STATE $cur\_pos = average\_distance * i$;
        \FOR{j=0; j $<$ N-1; j++}
        \IF{$len[j]\leqslant cur\_pos< len[j+1]$}
        \STATE $q_i=\frac{(p_{j+1}-p_j)*(cur\_pos-len[j])}{len[j+1]-len[j]} + p_j$;
        \ENDIF
        \ENDFOR
        \ENDFOR
    \end{algorithmic}
\end{algorithm}

\textbf{TotalText} contains horizontal, multi-oriented, and curved text in images. The dataset contains 1,255 training images and 300 test images. All images are annotated with polygons and transcriptions in word-level.

\textbf{ICDAR2015} focuses on multi-oriented scene text detection and recognition in natural images. There are 1,000 training images and 500 test images. Word-level quadrangles and transcriptions of each image are given.

\textbf{ICDAR2013} is a dataset which focuses on the horizontal scene text detection and recognition in natural images. The dataset consists of 229 images in the training set and 233 images in the test set. Similar to SynthText, the bounding box and the transcription for each word-level and character-level text instance are also provided.

\textbf{COCO-Text} contains 63,686 images. Though it is evaluated with axis-aligned bounding boxes, the text instances in the images are distributed in various orientations. Owing to no lexicon provided in the evaluation, the text must be recognized without any prior knowledge.

\subsection{Implementation Details}
Different from previous text spotting methods \cite{FOTS,ShenICCV2017} which use alternating training strategy, our boundary point detector and text recognizer could be trained in an end-to-end manner. The whole training process contains two steps: first, we pretrain the network on SynthText, then real data is adopted to finetune the model.

During pretraining, the mini-batch is set to 16, and the longer sides of input images are resized to 800 while keeping the aspect ratio. The maximum number of proposals in each image on the recognition branch is set to 16. In the finetuning stage, for the data augmentation, 
we randomly crop a patch whose edges range from 210 to 1100 while keeping all the text instance not cropped and resize the patch to (640, 640). 
Finally, the resized patch is randomly rotated $90^\circ$ with a probability of 0.2. We collect the training images from ICDAR2013, ICDAR2015, and TotalText to finetune the model with the mini-batch set to 16. 
We optimize our model using SGD with a weight decay of 0.0001 and momentum of 0.9. We train our model for 270k iterations for pretraining, with an initial learning rate of 0.01, and decayed to a tenth at the 100k and the 200k iteration. In the finetuning stage, the initial learning rate is set to 0.001 and then is decreased to 0.0001 and 0.00001 at the 80k and 120k iteration. The finetuning process is terminated at the 140k iteration.
We implement our method in Pytorch and conduct all experiments on a regular workstation with Nvidia Titan Xp GPUs. The model is trained in parallel and evaluated on a single GPU.

\textbf{Label Generation} During training, we need equidistantly spaced boundary points to train BPDN. However, only corner points are given in the groundtruth. So we need to sample points on the longer sides of text boundary using Algorithm \ref{alg:A}. In our experiments, $K$ is set to 7.

\begin{table}[tbp]
	\centering
	\begin{tabular}{|m{.32\columnwidth}<{\centering}|m{.07\columnwidth}<{\centering}m{.07\columnwidth}<{\centering}m{.07\columnwidth}<{\centering}|m{.07\columnwidth}<{\centering}m{.07\columnwidth}<{\centering}|}
		\hline
		\multirow{2}*{Method}  & \multicolumn{3}{c|}{Detection} & \multicolumn{2}{c|}{E2E} \\
		\cline{2-6} & P & R & F & None & Full \\
		\hline 
		\multirow{1}*{TotalText} & \multirow{2}*{40.0}    & \multirow{2}*{33.0}     & \multirow{2}*{36.0}   & \multirow{2}*{-} & \multirow{2}*{-} \\
		 \multirow{1}*{\small{\cite{ch2019total}}} &  &   &  &  &\\
		\hline 
		\multirow{1}*{TextBoxes} & \multirow{2}*{62.1}    & \multirow{2}*{45.5}     & \multirow{2}*{52.5}   & \multirow{2}*{36.3} & \multirow{2}*{48.9} \\
		 \multirow{1}*{\small{\cite{TextBoxes}}} &  &   &  &  &\\
		\hline
		\multirow{1}*{MaskTextSpotter} & \multirow{2}*{87.0} & \multirow{2}*{80.2} & \multirow{2}*{83.4} & \multirow{2}*{52.9} & \multirow{2}*{71.8} \\
		\multirow{1}*{\small{\cite{MaskTextspotter}}} &  &   &  &  & \\
		\hline
		\multirow{1}*{\textbf{Boundary} } & \multirow{2}*{85.2}  & \multirow{2}*{83.5} & \multirow{2}*{84.3} & \multirow{2}*{-}  & \multirow{2}*{-}  \\
		
		\multirow{1}*{\small{(det only)}} & & & & & \\
		    
		\hline
        \multirow{1}*{\textbf{Boundary}} & \multirow{2}*{\textbf{88.9}} & \multirow{2}*{\textbf{85.0}} & \multirow{2}*{\textbf{87.0}} & \multirow{2}*{\textbf{65.0}} & \multirow{2}*{\textbf{76.1}} \\
        \multirow{1}*{\small{(end-to-end)}}  & & & & & \\
		\hline
	\end{tabular}
	\caption{Results on TotalText. ``P", ``R" and ``F" mean Precision, Recall and F-measure in detection task respectively. ``E2E" means end-to-end, ``None" means recognition without any lexicon, ``Full" lexicon contains all words in test set. Following tables follow the same usage.}
	\label{totaltextresult}
\end{table}

\begin{table*}[t]
	\centering
	\begin{tabular}{|m{.65\columnwidth}<{\centering}|m{.06\columnwidth}<{\centering}m{.06\columnwidth}<{\centering}m{.06\columnwidth}<{\centering}|m{.06\columnwidth}<{\centering}m{.06\columnwidth}<{\centering}m{.06\columnwidth}<{\centering}|m{.06\columnwidth}<{\centering}m{.06\columnwidth}<{\centering}m{.06\columnwidth}<{\centering}|m{.06\columnwidth}<{\centering}m{.06\columnwidth}<{\centering}m{.06\columnwidth}<{\centering}|}
		\hline
		 & \multicolumn{6}{c|}{ICDAR2015} & \multicolumn{6}{c|}{ICDAR2013}\\
		\cline{2-13}
		\multirow{1}*{Method} & \multicolumn{3}{c|}{Detection} & \multicolumn{3}{c|}{E2E} & \multicolumn{3}{c|}{Detection} & \multicolumn{3}{c|}{E2E} \\
		\cline{2-13} & P & R & F  & S & W & G & P & R & F  & S & W & G \\
		\hline 
		DeepTextSpotter \small{\cite{busta2017deep}} & - & - & - & 54.0 & 51.0 & 47.0 & - & - & - & 89.0 & 86.0 & 77.0  \\
		\hline
		TextBoxes++ \small{\cite{liao2018textboxes++}} & 87.2 & 76.7 & 81.7 & 73.3 & 65.9 & 51.9 & 88.0 & 74.0 & 81.0 & 93.0 & 92.0 & 85.0 \\
		\hline
		He* $et\ al.$ \small{\cite{HeCVPR2018}}  & 87.0   & 86.0   & 87.0  & \textbf{82.0}  & \textbf{77.0}   & 63.0 & 91.0   & \textbf{89.0}    & 90.0 & 91.0 & 89.0 & 86.0 \\
		\hline
		FOTS* \small{\cite{FOTS}} & 91.0  & 85.2 & 88.0  & 81.1 & 75.9 & 60.8 & -  & - & 88.3 & 88.8 & 87.1 & 80.8          \\
		\hline
		MaskTextSpotter \small{\cite{MaskTextspotter}} & \textbf{91.6} & 81.0      & 86.0 & 79.3 & 73.0 & 62.4 & \textbf{95.0}   & 88.6   & \textbf{91.7} & \textbf{92.2} & \textbf{91.1} & \textbf{86.5}          \\
		\hline
		\textbf{Boundary} \small{(det only)}  & 88.1 & 82.2  & 85.0 & - & - & - & 89.3 &  85.2& 87.2& - & -& - \\
		\hline
		\textbf{Boundary} \small{(end-to-end)} & 89.8 & \textbf{87.5} & \textbf{88.6} & 79.7 & 75.2 & \textbf{64.1} & 93.1  & 87.3 & 90.1 & 88.2 & 87.7 & 84.1 \\
		\hline
	\end{tabular}
	\caption{Results on ICDAR2015 and ICDAR2013 (DetEval). ``S", ``W" and ``G" mean recognition with strong, weak and generic lexicon respectively. ``*" denotes that training dataset of MLT2017 is used for training. Following tables follow the same usage.}
	\label{icdarresult}
\end{table*}

\begin{table}[t]
	\centering
	\begin{tabular}{|m{.28\columnwidth}<{\centering}|m{.06\columnwidth}<{\centering}m{.06\columnwidth}<{\centering}m{.06\columnwidth}<{\centering}|m{.06\columnwidth}<{\centering}m{.06\columnwidth}<{\centering}m{.06\columnwidth}<{\centering}|}
		\hline
		\multirow{2}*{Method}  & \multicolumn{3}{c|}{Detection} & \multicolumn{3}{c|}{E2E}                                      \\
		\cline{2-7} & P & R & F & P & R & F    \\
		\hline
		 \multirow{1}*{Baseline A**} & \multirow{2}*{\textbf{83.8}}    & \multirow{2}*{23.3}     & \multirow{2}*{36.5}   & \multirow{2}*{\textbf{68.4}} & \multirow{2}*{28.3} & \multirow{2}*{40.0}   \\
		 \multirow{1}*{\small{\cite{COCOText}}} &  &   &  &  &  &\\
		
		\hline
		 \multirow{1}*{Baseline B**} & \multirow{2}*{\textbf{59.7}}    & \multirow{2}*{10.7}     & \multirow{2}*{19.1}   & \multirow{2}*{9.97} & \multirow{2}*{\textbf{54.5}} & \multirow{2}*{16.9}   \\
		 \multirow{1}*{\small{\cite{COCOText}}} &  &   &  &  &  &\\
		
		\hline
		\multirow{1}*{Baseline C**}       & \multirow{2}*{18.6}           & \multirow{2}*{4.7}                             & \multirow{2}*{7.5}           & \multirow{2}*{1.7}  & \multirow{2}*{4.2}  & \multirow{2}*{2.4}  \\
		\multirow{1}*{\small{\cite{COCOText}}} &  &   &  &  &  &\\
		
		\hline  
		\multirow{1}*{DeepTextSpotter}  & \multirow{2}*{-} & \multirow{2}*{-} &  \multirow{2}*{-} & \multirow{2}*{31.4} & \multirow{2}*{16.8} & \multirow{2}*{21.9} \\
		\multirow{1}*{\small{\cite{busta2017deep}}}  &  &   &  &  &  & \\
		
		\hline
		\multirow{1}*{\textbf{Boundary}}     & \multirow{2}*{59.0}       & \multirow{2}*{\textbf{67.7}}  & \multirow{2}*{\textbf{63.0}} & \multirow{2}*{55.7} & \multirow{2}*{32.8} & \multirow{2}*{\textbf{41.3}} \\
		\multirow{1}*{\small{(end-to-end)}} &  &   &  &  &  & \\
		\hline
	\end{tabular}
	\caption{Results on COCO-Text. Lexicon is not used for end-to-end testing. ``MS" means testing with multiple scales. Methods with ``**" are evaluated using V1.1 annotations. }
	\label{cocotextresult}
\end{table}

\subsection{Curved Text}
The proposed method focuses on arbitrary-shaped text spotting. To verify its effectiveness, we first conduct experiments on TotalText. During testing, the longer sides of images are resized to 1,100. For a fair comparison, we follow the evaluation protocols in the latest method \cite{MaskTextspotter}.

The performance on TotalText is given in Tab.~\ref{totaltextresult}. Our method achieves state-of-the-art performance on both detection and end-to-end recognition. Specifically, the proposed method outperforms MaskTextSpotter with improvements of 3.6\% and 12.1\% respectively on detection and end-to-end recognition without lexicon. The improvement over other methods gives credit to the following four points: 1) Compared with MaskTextSpotter, the attention-based decoder in our recognition module can capture the relationship between characters of a word, which is helpful for the recognition task. However, MaskTextSpotter predicts the characters separately, ignoring the context within them. 2) Compared with other methods, before being fed into the text recognizer, text with arbitrary shapes is rectified to a regular one, which attenuates text irregularities and therefore decreases the recognition difficulty. 3) Due to the better recognition results, the detection results could be implicitly improved through the shared backbone features. 4) Using boundary points to describe the shapes of text instances is more flexible and efficient to locate the text instances.

\subsection{Oriented Text}
We also conduct experiments on ICDAR2015 to confirm the superiority of the proposed method on the oriented scene text. Images are resized to 1080$\times$1920 before being fed into the framework. As shown in Tab.~\ref{icdarresult}, our method slightly outperforms previous methods in detection and end-to-end recognition with a general lexicon by 0.6\% and 1.1\% respectively. However, besides SynthText, extra 9,000 images in MLT2017 are used for training by \cite{FOTS,HeCVPR2018}. For a fair comparison, we follow the same settings with MaskTextSpotter, in which the images of MLT2017 are not used. We can observe that our method gets respectively 2.6\% and 1.7\% improvements in detection and end-to-end tasks with general lexicon.

\subsection{Horizontal Text}
Besides promising performances have been achieved on curved and oriented benchmarks, we also evaluate the proposed method on horizontal scene text. The longer sides of input images are resized to 1280 while keeping the aspect ratio of the images.
As shown in Tab.~\ref{icdarresult}, our method gets comparable performance on both tasks. However, character-level annotations are required for MaskTextSpotter. Additionly, extra 9,000 images from MLT2017 are added to improve the performance by FOTS and \cite{HeCVPR2018}.

\subsection{Generalization Evaluation}
We evaluate the generalization of our method on COCO-Text. Following \cite{lyu2018multi}, our model is not trained with the training set of COCO-Text. The detection task in Tab.~\ref{cocotextresult} is evaluated with the annotations V1.4 for a fair comparison with previous methods, while the end-to-end task is evaluated with the newest annotations V2.0.  The longer sides of input images are resized to 1280 while keeping the aspect ratio of images.
As shown in Tab.~\ref{cocotextresult}, our method achieves state-of-the-art performance on both tasks, which confirms that our method has stronger generalization ability.

\begin{table}[tb]
	\centering
\begin{tabular}{|m{.193\columnwidth}<{\centering}|c|m{.07\columnwidth}<{\centering}m{.07\columnwidth}<{\centering}m{.07\columnwidth}<{\centering}|m{.07\columnwidth}<{\centering}|}
		\hline \multirow{2}*{Datasets}  & \multirow{2}*{Proposal} & \multicolumn{3}{c|}{Detection} & {E2E}           \\
		\cline{3-6}  & & P & R & F    & None \\

		\hline \multirow{2}*{ICDAR2015}   &     axis-aligned      & 88.2 &\textbf{87.5} & 87.8 & 63.1    \\
		\cline{2-6} & oriented & \textbf{89.8} & \textbf{87.5} & \textbf{88.6} & \textbf{64.1}   \\
		\hline \multirow{2}*{TotalText} & axis-aligned & 87.3 & 84.3         & 85.8 & 63.8    \\
		\cline{2-6} & oriented & \textbf{88.9}  & \textbf{85.0}  & \textbf{87.0} & \textbf{65.0}   \\
		\hline
	\end{tabular}
	\caption{Comparison of axis-aligned proposal and oriented rectangular proposal. ``axis-aligned'' means that axis-aligned proposal is used to predict boundary points, ``oriented'' means that oriented rectangular proposal is used.}
	\label{table:abl1}
\end{table}

\begin{figure*}[ht]
\includegraphics[width=0.85\linewidth]{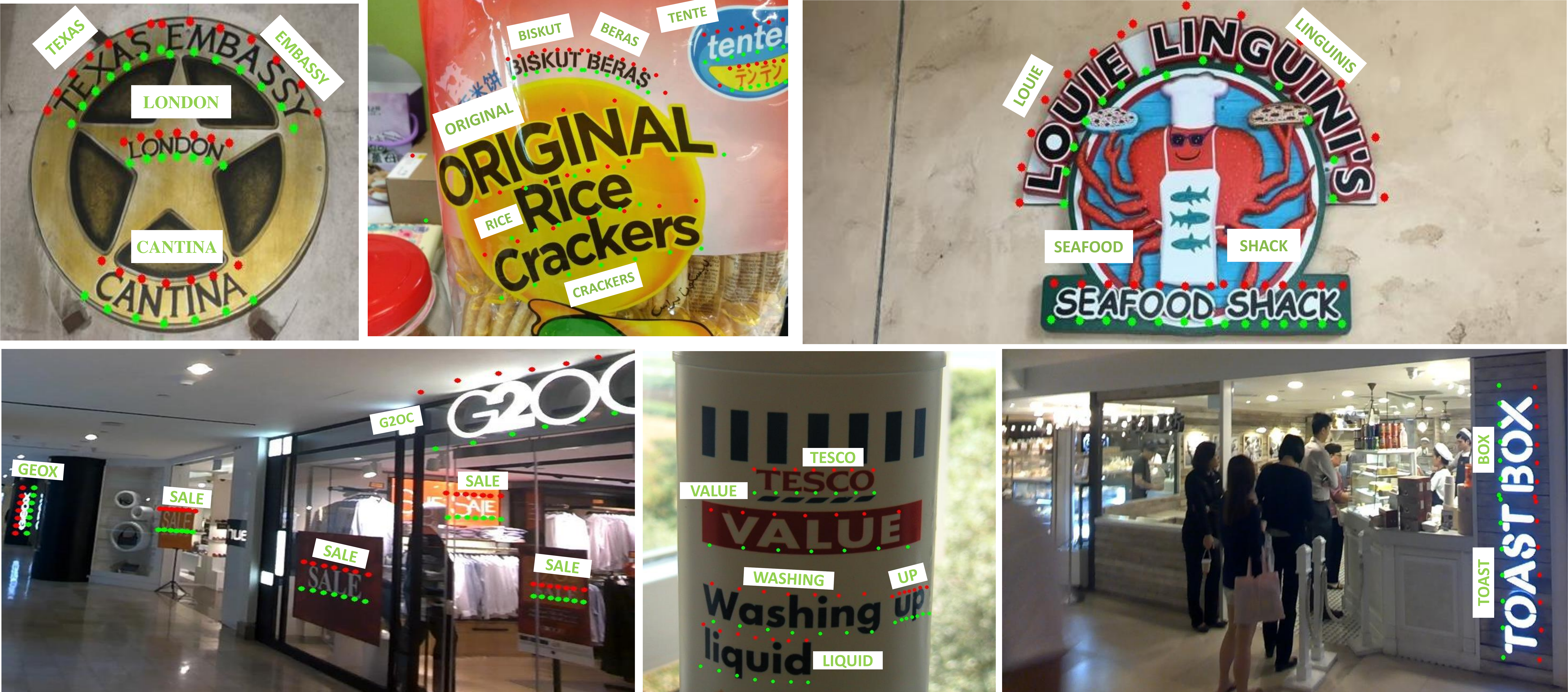}
\centering
  \caption{
  Examples of text spotting results of our method on Total-Text, ICDAR2015, and ICDAR2013. 
  }
  \label{result}
\end{figure*}

\begin{figure*}[ht]
\includegraphics[width=0.85\linewidth]{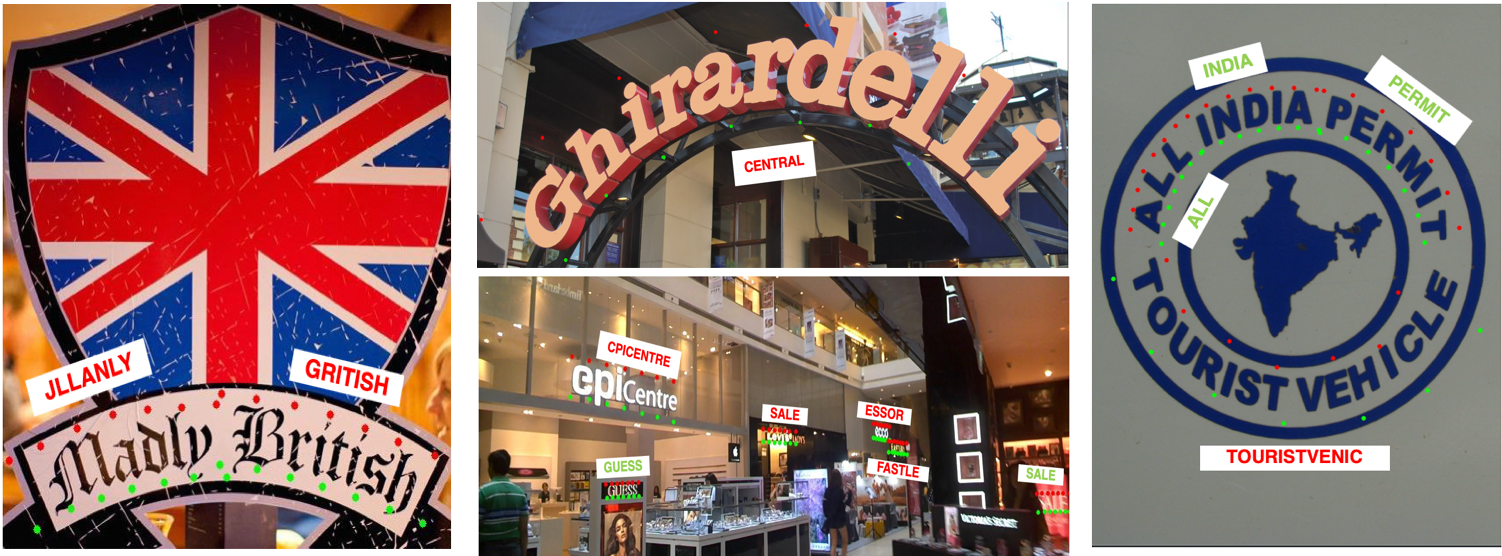}
\centering
  \caption{
  Some failure cases produced by our method.
  }
  \label{limitation}
\end{figure*}

\subsection{Visualization}
The results of several images containing text instances with arbitrary shapes are illustrated Fig.~\ref{result}. With the novel representation of text regions, the proposed system can read texts of arbitrary shape. Even for some vertical text instances (pictures at bottom left and bottom right of Fig.~\ref{result}), our method can successfully localize and recognize them. Those challenging samples confirm the superiority and robustness of the proposed boundary points. 

Although our method can achieve promising performances, there are still some failure cases, as shown in Fig. \ref{limitation}. We observe that it is difficult to correctly spot texts composed of rare art fonts, since there are few samples in the training set. Besides, our method also struggles to detect or recognize extremely tiny or long texts and blurred texts.

\subsection{Ablation Study}
\textbf{Oriented Rectangular Box Detector} As we mentioned before, the oriented rectangular box detector plays an important role in our whole pipeline. It provides better features through RotatedRoIAlign for boundary point regression. If we do not predict the oriented rectangular box and use the axis-aligned proposal from RPN, the boundary of text instance in the proposal changes dramatically, and it's hard to predict the precise location by regression. 

As shown in Tab. \ref{table:abl1}, using oriented rectangular proposal obtained from oriented rectangular box detector improves the performance of the detection and end-to-end recognition stably compared to the axis-aligned proposal. On ICDAR2015, the performance respectively improves by  0.8\% and 1.0\% in the detection and end-to-end tasks. On TotalText, the oriented rectangular proposal provides significant improvements of 1.2\% and 1.2\%. These results show that the oriented rectangular proposal can reduce the difficulty of boundary point prediction and make it more precise.

\section{Conclusion}
In this paper, we present an end-to-end trainable network that defines text of arbitrary shape as a set of boundary points.
Our method has achieved the state-of-the-art in both tasks of scene text detection and end-to-end text recognition on the standard benchmarks including oriented text and curved text, which confirms its robustness and effectiveness in reading scene text.
The flexible and accurate representation of boundary points is potential to become the mainstream description for scene text spotting.
Besides, both text detection and recognition tasks benefit from boundary point representation.
In the future, we would like to improve the efficiency of detecting boundary points.

\section{Acknowledgements}
This work was supported by National Natural Science Foundation of China (61733007, 61572207) and the Alibaba Innovative Research Program.

\bibliographystyle{aaai}
\bibliography{references}
\end{document}